\documentclass[letterpaper, 10 pt, conference]{ieeeconf}  

\IEEEoverridecommandlockouts                              

\usepackage{graphicx} 
\usepackage{amsmath} 
\usepackage{float} 
\usepackage{amssymb}
\usepackage{array}
\usepackage{booktabs} 
\usepackage{hyperref} 
\usepackage[all]{hypcap}
\usepackage{cite}
\usepackage{threeparttable}
\usepackage{xcolor} 

\title{\LARGE \bf
Learning-Based Passive Fault-Tolerant Control of a Quadrotor with Rotor Failure
}

\author{Jiehao Chen, Kaidong Zhao, Zihan Liu, YanJie Li*, Yunjiang Lou
\thanks{This work was supported by National Natural Science Foundation of China (61977019, U8131206) and Shenzhen Fundamental Research Program [JCYJ20180507183837726, JCYJ20220818102415033, JSGG2020103039802006]. (\textit{Corresponding author: Yanjie Li, autolyj@hit.edu.cn}.)} 
\thanks{The authors are with the Guangdong Key Laboratory of Intelligent Morphing Mechanisms and Adaptive Robotics and School of Inteligence Science and Engineering, the Harbin Institute of Technology Shenzhen, 518055, China.}
}

\begin{document}

\maketitle
\thispagestyle{empty}
\pagestyle{empty}

\begin{abstract}
This paper proposes a learning-based passive fault-tolerant control (PFTC) method for quadrotor capable of handling arbitrary single-rotor failures, including conditions ranging from fault-free to complete rotor failure, without requiring any rotor fault information or controller switching. Unlike existing methods that treat rotor faults as disturbances and rely on a single controller for multiple fault scenarios, our approach introduces a novel Selector-Controller network structure. This architecture integrates fault detection module and the controller into a unified policy network, effectively combining the adaptability to multiple fault scenarios of PFTC with the superior control performance of active fault-tolerant control (AFTC). To optimize performance, the policy network is trained using a hybrid framework that synergizes reinforcement learning (RL), behavior cloning (BC), and supervised learning with fault information. Extensive simulations and real-world experiments validate the proposed method, demonstrating significant improvements in fault response speed and position tracking performance compared to state-of-the-art PFTC and AFTC approaches. Video and code will be available at \url{https://github.com/HITSZcjh/uav_ftc}.
\end{abstract}

\section{INTRODUCTION}

As drones are increasingly applied across various industries, safety concerns have garnered significant attention. Among these concerns, rotor failures are particularly critical, often leading to the immediate crash of the drone. Compared to other multi-rotor UAVs, quadrotor systems lack actuator redundancy, making them especially vulnerable to rotor failures.

Fault-tolerant control (FTC) methods for addressing actuator failures aim to maintain the system's original control performance as effectively as possible in the presence of faults, without relying on additional hardware (e.g., parachutes). Generally, these methods are categorized into two types: active fault-tolerant control (AFTC) and passive fault-tolerant control (PFTC).

AFTC methods depend on a Fault Detection and Diagnosis (FDD) module to estimate fault information, which is subsequently used to either switch or reconfigure controllers for effective operation under faulty conditions \cite{zhang2008bibliographical}. In scenarios involving partial rotor failures, control performance can often be enhanced through thrust compensation \cite{liu2024reinforcement}. However, in cases of complete rotor failure, the quadrotor loses control of the yaw channel \cite{freddi2011feedback}, requiring a control strategy that is drastically different from fault-free conditions. Based on the concept of giving up the control of the yaw channel, numerous AFTC methods have been proposed,
\begin{figure}[]
    \centering
    \includegraphics[width=1\linewidth]{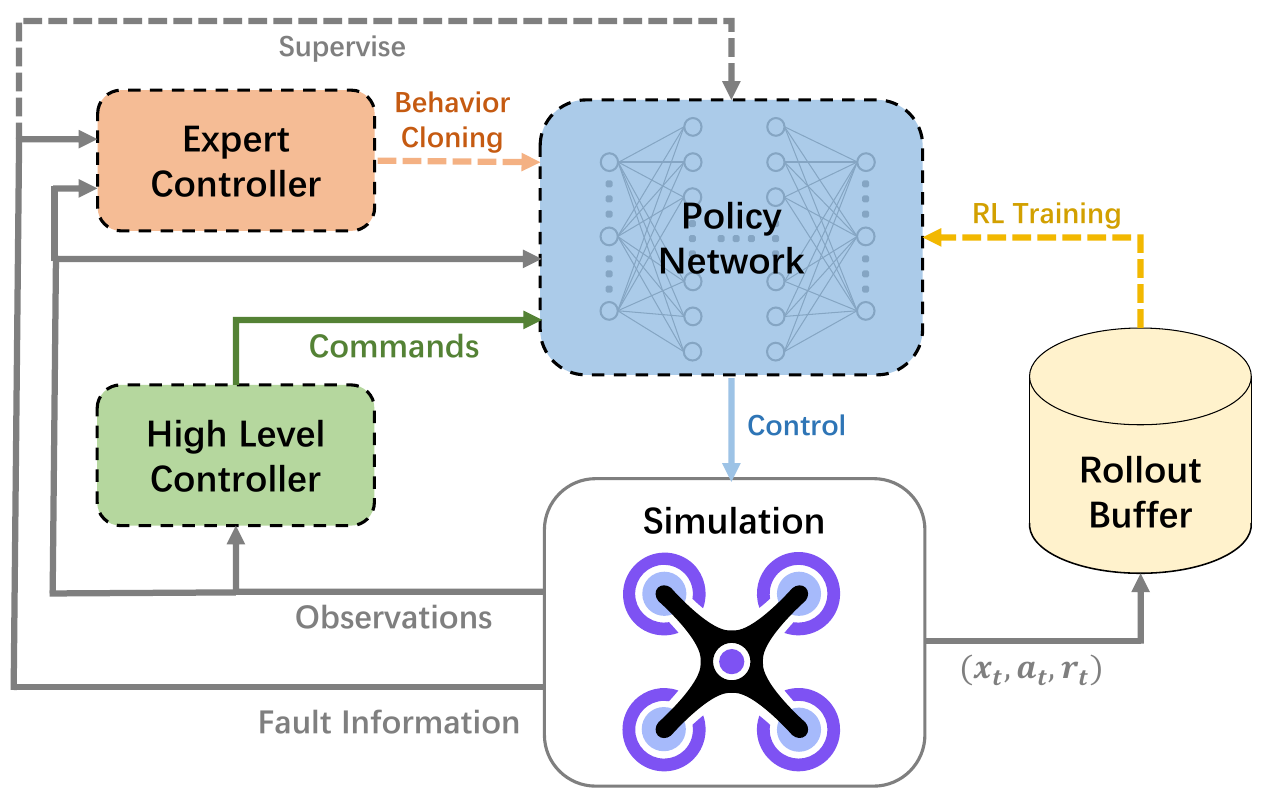}
    \caption{An overview of our policy training framework. The dashed lines represent updates to the policy parameters, while the solid lines indicate the flow of relevant variables. We train the policy using a combination of Reinforcement Learning (RL), Behavior Cloning (BC), and supervised learning with fault information.}
    \label{fig:total_structure}
\end{figure}
including both linearized methods\cite{lippiello2014emergency,mueller2014stability,stephan2018linear,de2015unified} and nonlinear methods\cite{sun2021autonomous,sun2018high,sun2020incremental,lippiello2014emergency_backstepping,nan2022nonlinear}. The linearized methods rely on the linearized dynamics of quadrotor, either around the hovering equilibrium point or along a predicted trajectory, to design controllers, which may degrade control performance. Notably, the work in \cite{mueller2014stability} showed that giving up the control of the yaw channel enables a quadrotor to remain controllable under single, double, or even triple rotor failures. The primary axis approach further simplifies stability analysis and controller design by decoupling tilt angle and primary axis control. In contrast, the nonlinear methods take into account the nonlinear dynamic model and significantly enhance control performance. The work in \cite{nan2022nonlinear,sun2018high} emphasize high-speed scenarios under rotor failure, demonstrating the quadrotor's performance limits. 

Most of the aforementioned AFTC methods utilize a cascade structure that combines a high-level controller with a low-level controller. The high-level controller computes the desired orientation of the primary axis based on the target position and then calculates the desired angular velocity. The low-level controller uses fault information along with high-level commands to control the quadrotor, ensuring accurate tracking of these commands. However, in the aforementioned studies, the experiments were conducted under ideal conditions, where fault information was known in advance, and the controller switch was manually executed simultaneously with the fault trigger. In practical applications, fault information must be obtained through an FDD module \cite{amoozgar2013experimental,freddi2010actuator}, which can introduce computational and observation delays as well as fault estimation errors. These factors can degrade the performance of AFTC methods and even lead to crash in severe rotor failure scenarios\cite{ke2023uniform}.
We also adopt the high-level and low-level cascade architecture but replaces the low-level controller with a policy network. The policy network takes high-level commands and the quadrotor's state as inputs, enabling both command tracking and fault inference, thereby eliminating the need for prior fault information. Moreover, since we use Reinforcement Learning (RL) to optimize the policy, the policy's objective is to maximize total rewards. Unlike the low-level controller in previous methods, which strictly tracks high-level commands, our approach provides greater flexibility and better meets more complex design requirements by designing various reward functions.

Unlike AFTC methods, PFTC methods integrate both fault-free and fault conditions into the controller design, making it possible to handle faults without switching controllers. This approach avoids the risk of performance degradation caused by cascading interactions between different modules in AFTC methods. Existing PFTC methods generally fall into two categories. The first focuses on scenarios involving partial rotor failures, where yaw channel is still controllable\cite{merheb2015design,barghandan2017improved}. The second addresses complete rotor failure scenarios\cite{beyer2022incremental,beyer2023incremental,ke2023uniform}. In such cases, the quadrotor's dynamics and control logic differ significantly from fault-free conditions. However, existing methods still treat rotor failure as a disturbance, which can lead to significant performance degradation. Notably, unlike the work in \cite{ke2023uniform}, which considers multiple rotor failure scenarios, we focus only on the scenarios involving the failure of a single rotor at any severity, based on the following considerations: First, Multiple rotor failures are rare and typically occur sequentially, allowing for timely detection and replacement \cite{liu2024reinforcement}. Second, scenarios with two adjacent or three rotor failures are controllable only when the rotor failures occur sequentially\cite{ke2023uniform,beyer2022incremental}. Moreover, multiple rotor failures can cause spin rates exceeding IMU measurement limits without special configurations\cite{ke2023uniform}. Therefore, control methods for multiple rotor failures are challenging to implement in the real-world applications and single-rotor failure scenarios are more realistic and practical to address.

All the aforementioned methods are model-based control methods, which can struggle in complex and dynamic situations such as complete rotor failure scenario. In contrast, learning-based control methods have gained significant attention in robotics for their ability to handle unknown scenarios and model uncertainties with exceptional adaptability \cite{kumar2021rma,loquercio2023learning,luo2024pie}. In UAV control, learning-based methods exhibits extreme adaptability, enabling a trained policy to effectively control quadrotor with significant parameter variations \cite{zhang2023learning}. This adaptability makes learning-based methods particularly suitable for complete rotor failure scenario, where the quadrotor's dynamics change drastically. However, most of the existing learning-based PFTC works only focus on partial rotor failure scenarios where the yaw channel remains controllable \cite{liu2024reinforcement,jiang2023active}, which are relatively less challenging.

Given the remarkable adaptability demonstrated by learning-based methods, we propose an learning-based PFTC method that aims to combine the advantages of existing PFTC and AFTC methods. This approach uses a unified policy network to avoid the instability associated with controller switching, while introducing a Selector-Controller network structure to effectively maintain optimal performance in each fault scenario. Overall, our work presents the following key contributions:
\begin{itemize}
\item We propose a learning-based PFTC method capable of achieving effective control under arbitrary single-rotor failure at any severity. To the best of our knowledge, we are the first to employ learning-based methods for real-world quadrotor validation under complete rotor failure conditions.

\item Our method has been validated through both simulation and real-world experiments, demonstrating significant improvements in fault response speed and position tracking performance compared to state-of-the-art PFTC methods \cite{ke2023uniform} and leading AFTC methods \cite{nan2022nonlinear, sun2018high}. Furthermore, our method achieves performance levels comparable to AFTC methods that use prior fault information.

\item We introduce a Selector-Controller network architecture to combine the advantages of existing PFTC and AFTC methods and update the policy network using a combination of RL, Behavior Cloning (BC), and supervised learning with fault information, which further enhances the system's control performance.
\end{itemize}
\begin{figure}[b]
    \centering
    \includegraphics[width=0.3\textwidth]{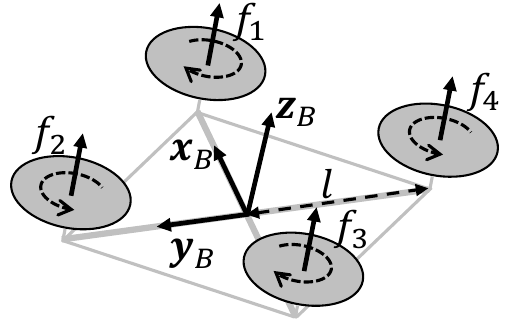}
    \caption{Definition of quadrotor body frame and rotor indices.}
    \label{fig:uav}
\end{figure}
\section{METHODOLOGY}
The following section outlines the quadrotor modeling methodology subject to rotor failure and presents our learning-based PFTC design. Notation conventions are defined as follows: bold lowercase letters (e.g., $\boldsymbol{v}$) denote vectors, bold uppercase letters (e.g., $\boldsymbol{M}$) denote matrices, and all other symbols represent scalars. We adopt two right-handed coordinate systems: the body frame ${\boldsymbol{x}_B, \boldsymbol{y}_B, \boldsymbol{z}_B}$ and the inertial frame ${\boldsymbol{x}_I, \boldsymbol{y}_I, \boldsymbol{z}_I}$. The body frame is defined as is shown in Fig.~\ref{fig:uav} while the inertial frame is defined with $\boldsymbol{z}_I$ pointing upward opposite to gravity. Quantities expressed in the inertial frame are superscripted with $I$ (e.g., $\boldsymbol{a}^I$), while those in the body frame are superscripted with $B$ (e.g., $\boldsymbol{a}^B$).
\subsection{Quadrotor Modeling Subject to Rotor Failure}
\label{sec:model}
As illustrated in the Fig.~\ref{fig:uav}, the thrusts generated by the four rotors are denoted as $f_1$ to $f_4$, and $l$ represents the distance from the quadrotor's center of mass (CoM) to each rotor. Since a symmetric frame is adopted, it is assumed that all four rotors are equidistant from the CoM. Based on these definitions, the dynamics of the quadrotor can be expressed as follows \cite{nan2022nonlinear}:
\begin{equation}
\begin{array}{ll}
\dot{\boldsymbol{p}} = \boldsymbol{v}, & \dot{\boldsymbol{v}} = \frac{1}{m}(\boldsymbol{R}\boldsymbol{(q)} 
\begin{bmatrix}
0 \\
0 \\
F
\end{bmatrix} 
+ \boldsymbol{F}_a) + \boldsymbol{g}, \\
\dot{\boldsymbol{q}} = \frac{1}{2} \boldsymbol{q} \circ 
\begin{bmatrix}
0 \\
\boldsymbol{\omega}
\end{bmatrix}, &
\dot{\boldsymbol{\omega}} = \boldsymbol{J}^{-1} 
\left( \boldsymbol{\tau} - \boldsymbol{\omega} \times \left( \boldsymbol{J} \boldsymbol{\omega} \right) + \boldsymbol{\tau}_{a} \right).
\end{array}
\end{equation}
where $\boldsymbol{p}$ and $\boldsymbol{v}$ represent the position and velocity of the quadrotor in the inertial frame, respectively. $\boldsymbol{\omega} = \begin{bmatrix} \omega_x & \omega_y & \omega_z \end{bmatrix}^T$ represents the angular velocity in the body frame and $\boldsymbol{q}$ denotes the rotation quaternion from the body frame to the inertial frame. 
$\boldsymbol{R}\boldsymbol{(q)}$ denotes the rotation matrix corresponding to $\boldsymbol{q}$. $\boldsymbol{J}$ is the moment-of-inertia matrix of the quadrotor, $m$ is the mass of the quadrotor, and $\boldsymbol{g}$ is the gravitational acceleration vector.
The $\circ$ operator denotes the quaternion multiplication. $\boldsymbol{F}_a$ and $\boldsymbol{\tau}_a$ denote the aerodynamic drag force and torque, respectively, while $F$ and $\boldsymbol{\tau}$ represent the total thrust and torque generated by the rotors. The relationship between $F$, $\boldsymbol{\tau}$, and the individual rotor thrusts $\boldsymbol{f} = \begin{bmatrix} f_1 & f_2 & f_3 & f_4 \end{bmatrix}^T$ is expressed as follows:
\begin{equation}
\begin{bmatrix}
F \\
\boldsymbol{\tau}
\end{bmatrix}
= \begin{bmatrix}
1 & 1 & 1 & 1 \\
0 & l & 0 & -l \\
-l & 0 & l & 0 \\
k_m & -k_m & k_m & -k_m
\end{bmatrix}
\boldsymbol{f}
\end{equation}
where $k_m$ represents the torque coefficient. For the commanded thrusts $\boldsymbol{u} = \begin{bmatrix} u_1 & u_2 & u_3 & u_4 \end{bmatrix}^T$ and the rotor thrust $\boldsymbol{f}$ , the relationship can be modeled using a first-order inertial system, similar to \cite{nan2022nonlinear}, as follows
\begin{equation}
\dot{{f}}_i=\frac{1}{\tau_f}\left(k_iu_i-f_i\right), \quad i=1,2,3,4
\end{equation}
where $\tau_f$ denotes the time constants of the motor and $\boldsymbol{k} = \begin{bmatrix} k_1 & k_2 & k_3 & k_4 \end{bmatrix}^T$ denotes the rotor's failure coefficient; $k_i \in [0,1]$; $k_i=0$ indicates that the $i$th rotor is completely failed and $k_i=1$ indicates that the $i$th rotor is fault-free.
\begin{figure}[b]
    \centering
    \includegraphics[width=1\linewidth]{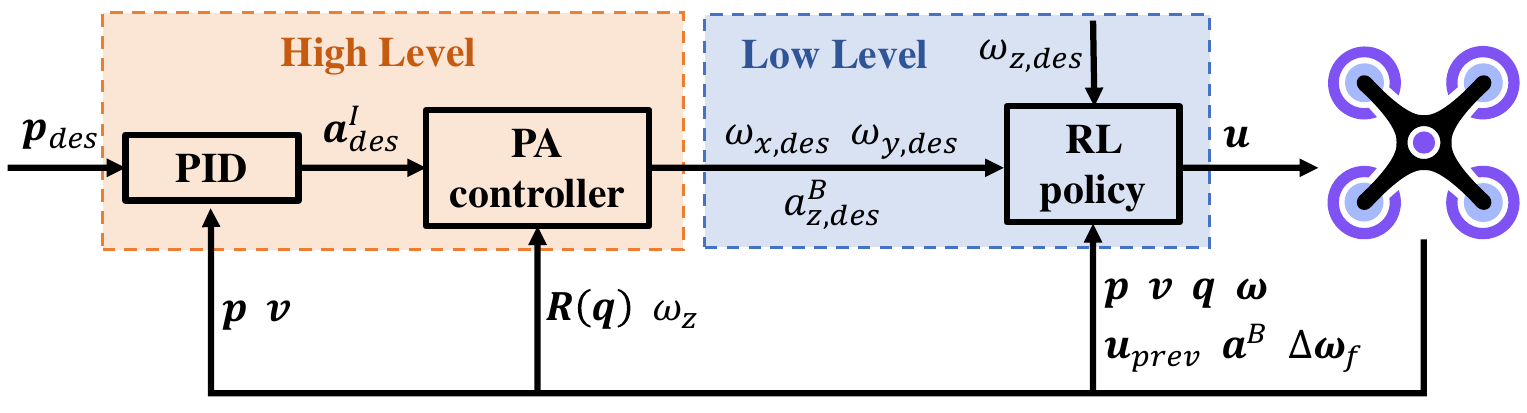}
    \caption{Control structure of high-level controller and low-level controller.}
    \label{fig:pa_controller}
\end{figure}
\subsection{Learning-Based Passive Fault Tolerant Controller}
Fig.~\ref{fig:pa_controller} shows the detailed control framework, where the subscript $des$ represents desired values, $\boldsymbol{a}$ is the quadrotor's acceleration and $\Delta\boldsymbol{\omega}_f$ denotes the angular acceleration approximated by numerical differentiation and filtering.

Similar to many AFTC methods, we use a Proportional-Derivative-Integral (PID) position controller cascaded with a Primary-Axis (PA) attitude controller as the high-level controller. This setup enables effective position control and primary axis management without the need for yaw torque control. For the low-level controller, we employ an RL-based policy. The policy takes high-level commands and the quadrotor’s state as inputs, enabling both command tracking and fault inference. Since the design of the low-level controller is the focus of this paper and the high-level controller design is already well-established with numerous existing implementations\cite{sun2021autonomous,sun2018high,sun2020incremental}, we omit the details of the high-level controller design and discuss the design of the policy as follows:

\subsubsection{Observation and Action Space}
The observation vector of the policy network, $\boldsymbol{x} \in \mathbb{R}^{27}$, includes the quadrotor's position $\boldsymbol{p}$, velocity $\boldsymbol{v}$, attitude $\boldsymbol{q}$, angular velocity $\boldsymbol{\omega}$, last step policy output $\boldsymbol{u}_{prev}$, desired angular velocity $\boldsymbol{\omega}_{des} = \begin{bmatrix} \omega_{x,des}&\omega_{y,des}&\omega_{z,des}\end{bmatrix}^T$, and desired $\boldsymbol{z}_B$ axis acceleration $a^B_{z,des}$. Notably, $\omega_{x,des}$, $\omega_{y,des}$ and $a^B_{z,des}$ are the high-level commands and $\omega_{z,des}$ is the manual control command. To effectively infer fault information, it also incorporates IMU-measured acceleration $\boldsymbol{a}$ and approximated angular acceleration $\Delta \boldsymbol{\omega}_f$, as rotor failures cause a mismatch between commanded thrusts $\boldsymbol{u}$ and actual thrusts $\boldsymbol{f}$, which directly influence angular and linear accelerations. The action vector of the policy network, $\boldsymbol{u} \in \mathbb{R}^{4}$, is the commanded thrusts.
\begin{figure}[t]
\vspace{5pt} 
    \centering
    \includegraphics[width=1\linewidth]{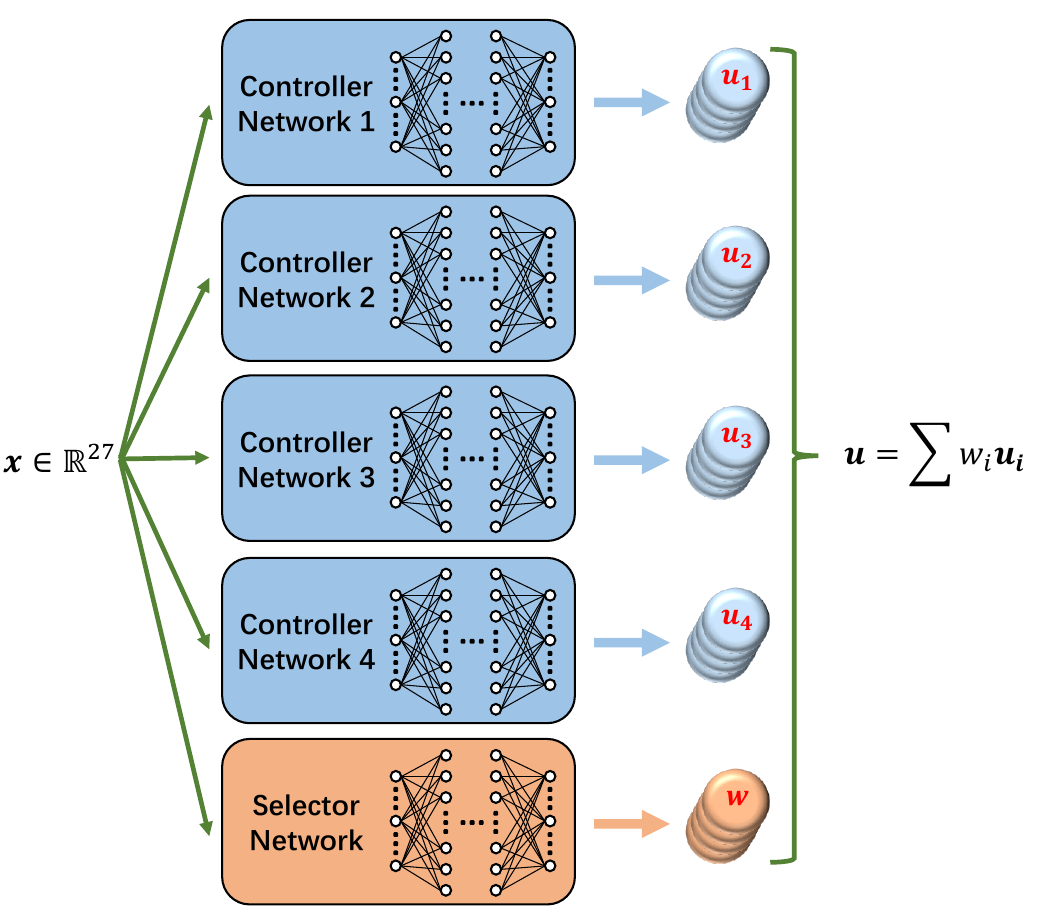}
    \caption{The Selector-Controller network structure consists of identical MLP architectures for both the selector and controller networks. The final control output is computed as a weighted sum of the outputs from all controller networks, where the weights are determined by the selector network.}
    \label{fig:network}
\end{figure}
\subsubsection{Selector-Controller Network}
Existing PFTC methods treat rotor failure as a disturbance and  use a single controller for all fault scenarios, which often face significant performance losses. In contrast, AFTC methods employ fault-scenario-specific controllers to optimize performance for each fault case. However, AFTC methods rely on the FDD module, which can introduce delays and estimation errors, affecting overall performance.
To address the limitations of existing PFTC and AFTC methods, we designed a Selector-Controller network architecture that combines their strengths. As shown in Fig.~\ref{fig:network}, we designed four identical controller networks, each targeting one of the four rotor failure scenarios, along with a selector network. The final control output is a weighted sum of the outputs of all controller networks, with the weights calculated by the selector network. 

If the selector network outputs constant weights at all fault scenarios, the proposed network is the same as a single Multi-Layer Perceptron (MLP) network. In this case, the gradients back-propagated during training for different failure scenarios may conflict, resulting in a performance trade-offs. Conversely, if the selector outputs ideal fault labels (e.g., $[1, 0, 0, 0]$ when the first rotor fails), the policy switches to the corresponding controller, avoiding gradient conflicts and maintaining optimal performance for each scenario.

Notably, unlike AFTC methods, which separately design FDD modules and controllers, our approach jointly optimizes the selector and controller networks. This integration allows controllers to account for issues such as observation delays and fault estimation errors introduced by the selector, resulting in improved robustness compared to AFTC methods. The selector supervise loss minimizes the mean squared error loss on the selector outputs and labels:
\begin{equation}
L_{selector}(\pi)=\left\|\boldsymbol{w}_{{label}}-\boldsymbol{w}\right\|^2
\end{equation}

\subsubsection{Rewards Design}
The reward function guides the agent to approach and hover at the goal position while penalizing oscillations under single rotor failures of any severity. Specifically, for partial rotor failures (assumed as $k_i > 0.5$ in our experiments) where yaw channel is controllable, the agent is encouraged to avoid spinning. For cases yaw channel is uncontrollable, the reward aims to limit spinning angular velocity to avoid exceeding the IMU measurement range:
\begin{itemize}
\item Position tracking penalty: $-\|\boldsymbol{p} - \boldsymbol{p}_{des}\|^2$  
\item Velocity penalty: $-\|\boldsymbol{v}\|^2$  
\item Angular velocity penalty (yaw channel controllable):  
\[
- \|\boldsymbol{\boldsymbol{\omega}} - \boldsymbol{\omega}_{des}\|^2
\]
Angular velocity penalty (yaw channel uncontrollable):  
\[
-(\omega_x - \omega_{x,des})^2 - (\omega_y - \omega_{y,des})^2 - H(|\omega_z| - C)(|\omega_z| - C)^2
\]
where $y=H(x)$ is the unit step function.
\item Oscillation penalty: $-\|\boldsymbol{u}_t - \boldsymbol{u}_{t-1}\|^2$  
\item Acceleration tracking penalty: $-(a_z^B - a_{z,des}^B)^2$
\end{itemize}

The reward is a weighted sum of above terms, with weights of $0.1$, $0.01$, $0.02$, $0.0008$, and $0.2$, respectively. We then optimize the expected return of the policy $\pi$ using Proximal Policy Optimization (PPO) method to compute the policy loss, denoted as $L_{RL}(\pi)$.

\subsubsection{Behavior Cloning}
To optimize control performance, we incorporate AFTC expert actions to guide the training process, where the expert controller uses prior fault information, whereas the policy does not.

When yaw channel is controllable, we use the PID control method to obtain the desired values $\boldsymbol{\omega}_{des}$ and $a_{z,des}^B$. In yaw channel uncontrollable scenarios, we use PA control attitude controller\cite{sun2018high} to obtain $\omega_{x,des}$, $\omega_{y,des}$, and $a_{z,des}^B$. Desired torque is then calculated using feedback linearization methods:
\begin{equation}
\boldsymbol{\tau}_{des} = -\boldsymbol{J}\boldsymbol{K} (\boldsymbol{\omega} - \boldsymbol{\omega}_{des}) - \boldsymbol{\tau}_a + \boldsymbol{\omega} \times (\boldsymbol{J}\boldsymbol{\omega})
\end{equation}
where $\boldsymbol{K}$ is the gain matrix. With the $\boldsymbol{\tau}_{des}$, $a_{z,des}^B$ and  rotor’s failure coefficient $\boldsymbol{k}$, it's easy to calculate the desired rotor thrusts $\boldsymbol{u}_{des}$ and the behavior cloning loss is expressed as follow: 
\begin{equation}
L_{BC}(\pi)=\|\boldsymbol{u}_{des}-\boldsymbol{u}\|^2
\end{equation}

Finally, as shown in Fig.~\ref{fig:total_structure}, we update the policy with a combinded loss as follow:
\begin{align}
L(\pi) &= w L_{selector}(\pi) + (1-\alpha)L_{RL}(\pi) + \alpha L_{BC}(\pi) \\
\alpha &= e^{-0.01t_{\text{epoch}}}
\end{align}
where $w$ is a weight factor and $\alpha$ is a decay factor.

\begin{table}[]
\vspace{5pt} 
\caption{Quadrotor Parameters}
\centering
\begin{tabular}{ccc}
\toprule  
Parameter & Description & Values \\
\midrule  
$m$ (kg) & Quadrotor mass & 0.764 \\
$\boldsymbol{J}$ (kg·m$^2$) & Moment of inertia & diag(0.0036, 0.0029, 0.0053) \\
$l$ (m) & Arm length & 0.125 \\
$\tau_f$ (s) & Rotor time constant & 0.025 \\
\bottomrule 
\end{tabular}
\label{table:params}
\end{table}

\section{EXPERIMENTAL SETUP}
\subsection{Simulation}
We use the Flightmare simulator\cite{song2020flightmare} for training and testing our control policies. The time step is discretized to 0.02 seconds, with a maximum of 500 steps per episode. However, for real-world deployment, the policy operates at a frequency of 400 Hz. The training task is designed for hover stabilization, with the goal position $\boldsymbol{p}_{des} = [0, 0, 3]^T$. Initially, the quadrotor is randomly generated in a fault-free state within a cubic region of 1-meter side length around the goal point. At step 250, a random severity of fault is injected into one of the rotors. 
\begin{figure*}
\vspace{5pt} 
    \centering
    \includegraphics[width=1\linewidth]{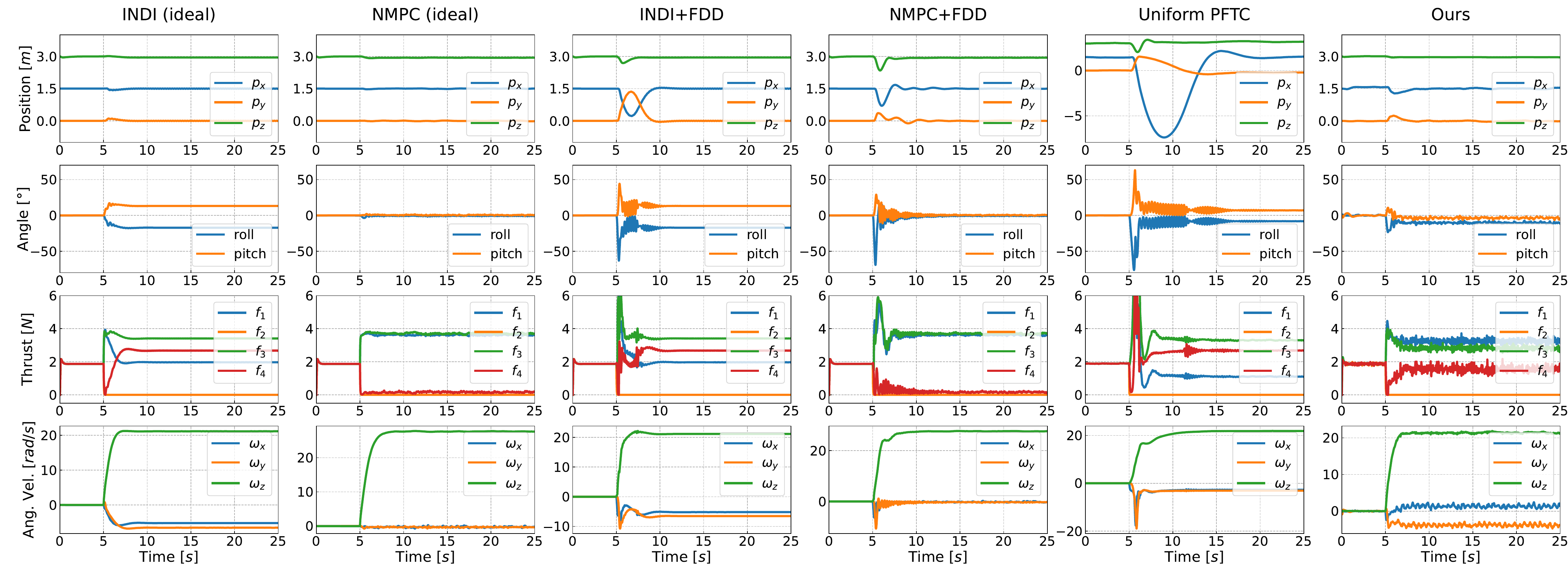}
    \caption{Time history of the quadrotor states controlled by various methods in the simulations. \textbf{``INDI''} and \textbf{``NMPC''} methods are the AFTC methods. \textbf{``ideal''} represents a controller switch was performed simultaneously with the fault trigger and \textbf{``FDD''} represents that an FDD module was conducted to detect faults and switch controllers as needed. \textbf{``Uniform PFTC''} and \textbf{``Ours''} methods are the PFTC methods, which do not require a controller switch or prior knowledge of fault information.}
    \label{fig:recover}
\end{figure*}
\begin{figure}
    \centering
    \includegraphics[width=1\linewidth]{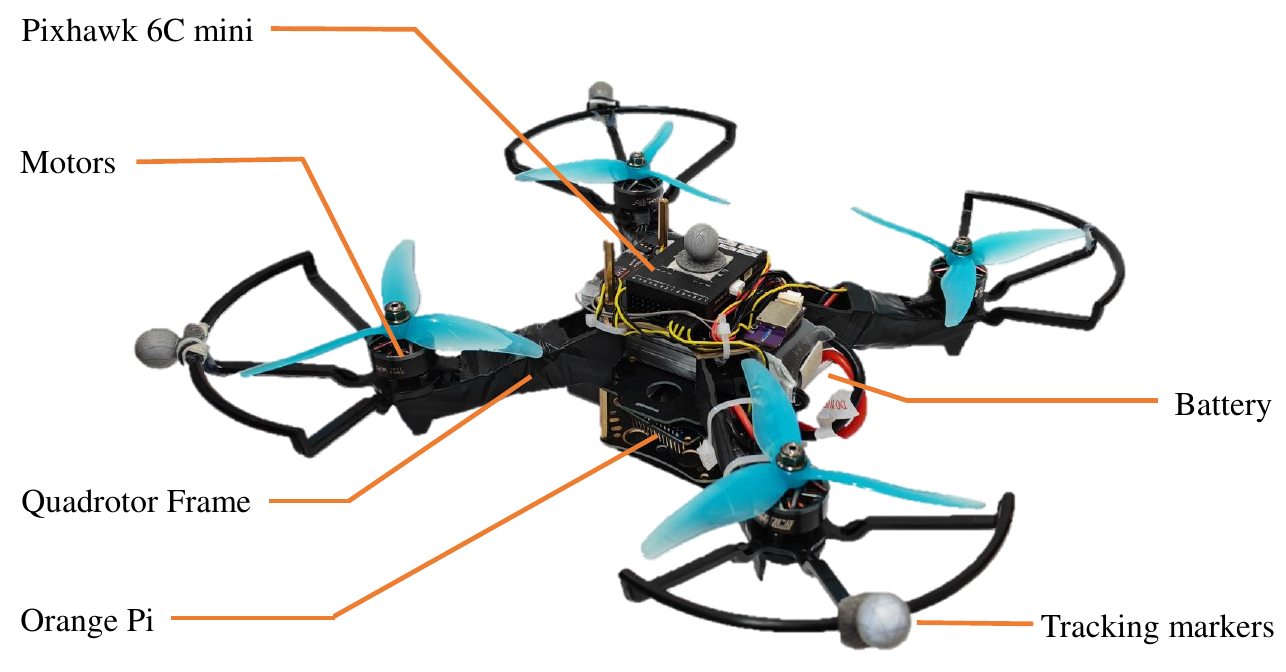}
    \caption{The quadrotor used in the real-world experiment.}
    \label{fig:hardware}
\end{figure}
\subsection{Hardware Details}
We validated our algorithm using a Q250 quadrotor frame equipped with T-MOTOR F60PRO motors paired with 5-inch propellers, capable of generating a maximum thrust of $13N$ per rotor. The relevant parameters of the quadrotor are summarized in Table~\ref{table:params}. The position and attitude of the quadrotor are observed using an external motion capture system, which transmits data to the onboard computer (OrangePi) via WiFi at a frequency of 100 Hz. The onboard computer then relays this information to the PX4 flight controller through MAVROS. The Extended Kalman Filter deployed on PX4 integrates the data to obtain reliable position, velocity, and attitude estimates. Meanwhile, angular velocity and linear acceleration are directly acquired from the onboard IMU. Fig.~\ref{fig:hardware} presents the quadrotor diagram for real-world experiments.

\subsection{Network Structure} 
To ensure the deployability of our algorithm on PX4 flight controllers, we designed lightweight network architectures. Each controller network and selector network is a 3-layer MLP network with 64 neurons in each hidden layer. The runtime of the policy on the STM32H7 flight controller is approximately 0.8 ms, which satisfies the requirements for real-time operation. In addition, we directly deploy the same controller trained in the simulation without any modifications. The policies was trained using the PPO algorithm over 1.2 billion steps. The entire training process took approximately 6 hours on a standard desktop equipped with an RTX4060-Laptop GPU.
\begin{figure}
    \centering
    \includegraphics[width=1\linewidth]{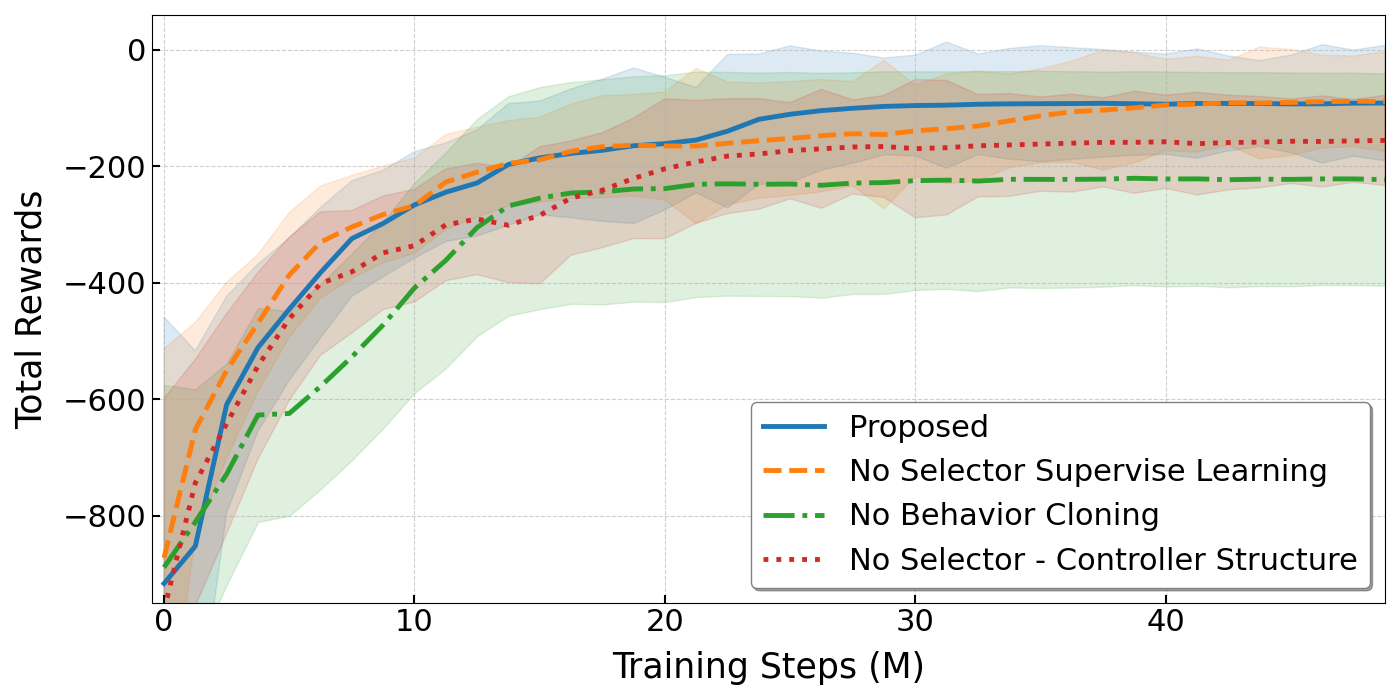}
    \caption{Total rewards curves from the training process under various configurations. \textbf{``No Selector Supervise Learning''} represents $L_{selector}(\pi)=0$, \textbf{``No Behavior Cloning''} represents $L_{BC}(\pi)=0$, \textbf{``No Selector-Controller Structure''} indicates replacing the Selector-Controller network with a single MLP network of equivalent parameter size.}
    \label{fig:returns}
\end{figure}

\section{RESULT}
\subsection{Simulation Result}
\subsubsection{Ablation Experiments}
\begin{figure*}
\vspace{5pt} 
    \centering
    \includegraphics[width=1\linewidth]{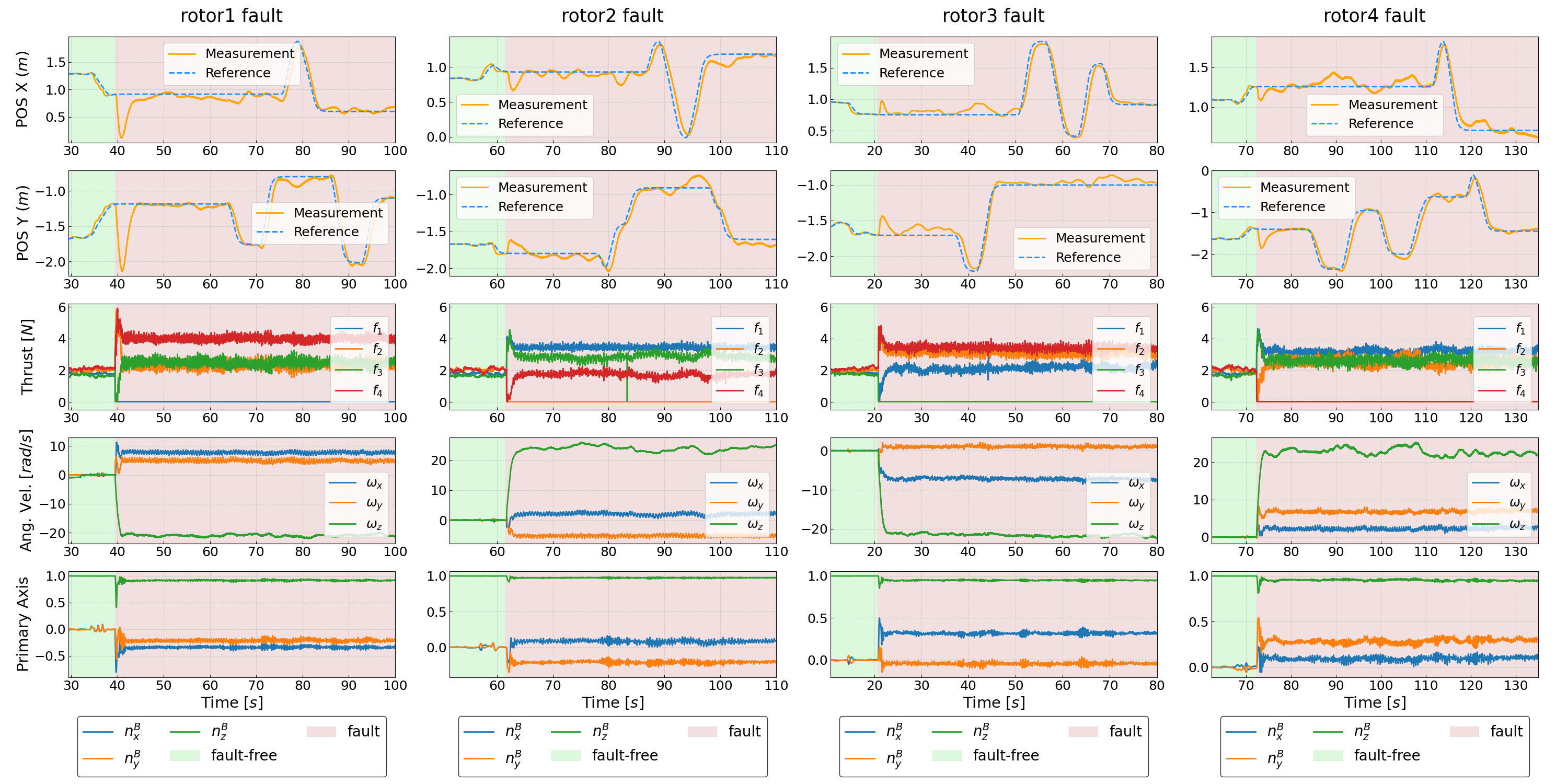}
    \caption{Time history of the quadrotor states controlled by the proposed method in the real-world experiment under complete rotor failure scenarios.}
    \label{fig:real_0}
\end{figure*}
To demonstrate the advantages of the proposed learning-based training framework, we conducted ablation experiments, as shown in the Fig.~\ref{fig:returns}. Compared to configurations without the Selector-Controller network structure, the proposed method achieves faster convergence and superior final performance. This is because the Selector-Controller network structure effectively avoids gradient conflicts in different failure scenarios. Additionally, compared to configurations without behavior cloning, the proposed method achieves superior performance because the AFTC expert helps guide the policy to learn more effectively. Additionally, the presence or absence of Selector Supervise Learning has minimal impact on the training outcomes. This is because, even without Selector supervise learning, the parameters of the Selector network are updated through RL training and behavior cloning, enabling the Selector to learn to extract fault information from the quadrotor states and effectively switch the controller in different fault scenarios.

\subsubsection{Recovery From Complete Rotor Failure}
To validate that our method enhances the quadrotor's performance compared to existing PFTC methods and AFTC methods, we conduct a recovery experiment from complete rotor failure with various methods. The quadrotor was initially controlled to hover at a reference position $(1.5,0,3)$ under fault-free conditions. At $t=5s$, a complete rotor failure occurred in the second rotor, and the reference position remained unchanged from the fault-free condition. The comparisons included two state-of-the-art AFTC methods: NMPC \cite{nan2022nonlinear} and INDI \cite{sun2018high}, as well as the leading PFTC approach, Uniform PFTC \cite{ke2023uniform}. 

For the two AFTC methods, an Extended Kalman Filter (EKF)-based FDD module was used to detect faults and switch controllers as needed. In contrast, neither our method nor the Uniform PFTC method requires a controller switch or prior fault information. Additionally, experiments under ideal conditions with the AFTC methods were also conducted, where a controller switch was performed simultaneously with the fault trigger.

As shown in Fig.~\ref{fig:recover}, compared to AFTC methods under ideal conditions, the AFTC+FDD methods exhibit larger roll and pitch angles during fault occurrence, which can further lead to greater position fluctuations. This performance degradation is mainly due to the approximately 0.1s observation delay introduced by the FDD module during the experiment. In contrast, our method does not require controller switching and can respond to faults promptly, resulting in superior performance compared to the AFTC+FDD approach. Additionally, when compared to AFTC methods under ideal conditions, our method demonstrates comparable control performance.


Compared to the Uniform PFTC method, which treats rotor failure as a disturbance, our approach uses the Selector-Controller network structure to integrate the fault detection module and the controller into a unified policy network. This structure combines the advantages of AFTC methods in isolating different fault scenarios with the benefits of PFTC methods in using a single controller, resulting in faster fault response and smaller position fluctuations, as shown in Fig.~\ref{fig:recover}.


\begin{figure}
    \centering
    \includegraphics[width=1\linewidth]{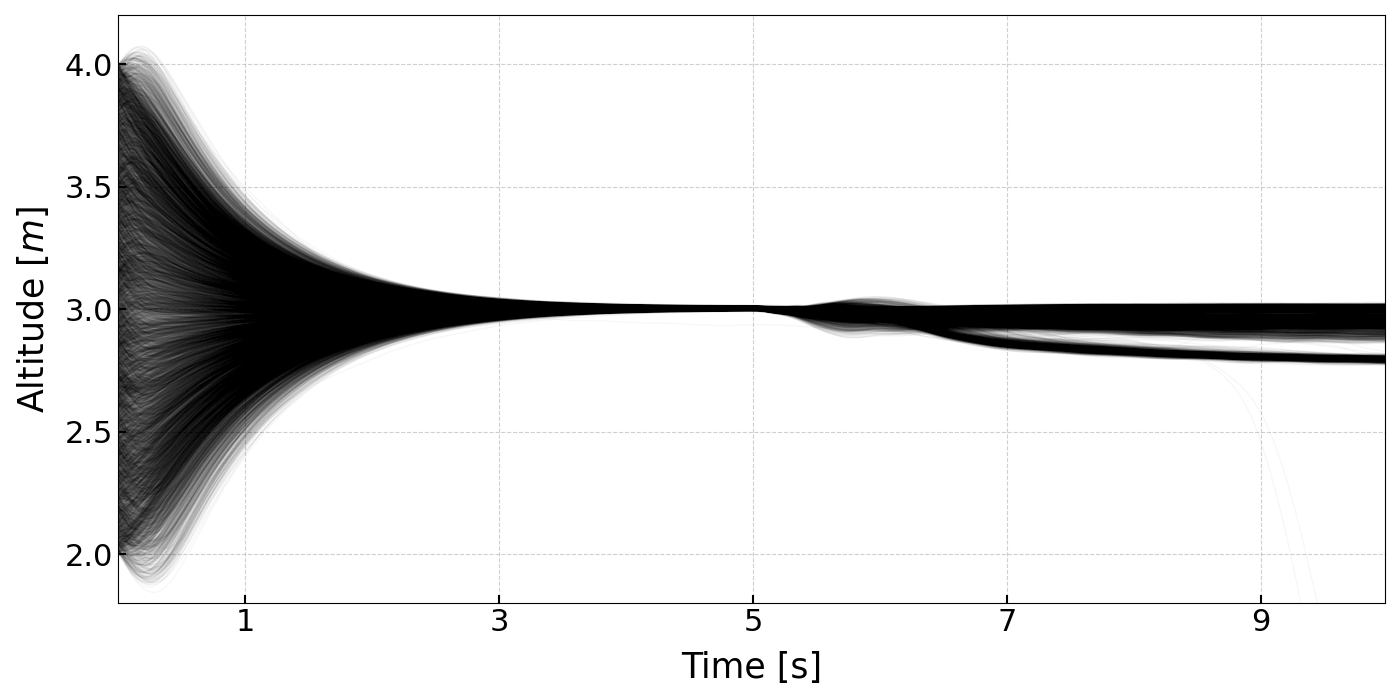}
    \caption{Time history of the quadrotor altitude controlled by the proposed method during recovery from rotor failure at randomized severities.}
    \label{fig:recover_all}
\end{figure}
\subsubsection{Recovery from Rotor Failure at Any Severity}
To validate that our method can achieve effective control under arbitrary single-rotor failures at any severity, we conducted extensive randomized simulations. In these simulations, the quadrotor was initialized at various positions and attitudes under fault-free conditions, and a random single-rotor failure with random severity was triggered at $t=5s$. After the fault, the quadrotor was controlled to hover at the same position. The simulation was performed 10,000 times, and the altitude curves over time for these flights are shown in Fig.~\ref{fig:recover_all}. In most of the flight tests, our method effectively controlled the quadrotor. It showed excellent position
tracking performance under fault-free conditions as well as strong adaptability to various
fault scenarios.

\subsection{Real World Experiment}
To evaluate the practical effectiveness of our algorithm, we conducted experiments using a real-world quadrotor and compared our method against the AFTC method, NMPC \cite{nan2022nonlinear}, the Uniform PFTC method \cite{ke2023uniform}, and the Incremental PFTC method \cite{beyer2023incremental}. For the comparison, we directly matched the experimental data from our method with the data reported in their respective papers. Following the experimental setup outlined in \cite{ke2023uniform}, the task was a position-tracking mission under complete rotor failure scenarios, with position control commands sent through a remote controller. Notably, the experiment for the AFTC methods was conducted under ideal conditions, where the controller switch was manually executed simultaneously with the fault trigger.

As shown in the position curves in Fig.~\ref{fig:real_0}, our method effectively overcame the quadrotor instability caused by rotor failure and achieved precise position control after the failure occurred. Table\ref{table:experiment} compares our method with three baseline approaches, evaluating metrics including the maximum $x$-$y$ position error, the average $x$-$y$ position error, and the average $x$-$y$ velocity error. As shown in Table~\ref{table:experiment}, our method delivers significant position tracking performance improvements over the two PFTC methods and achieves results comparable to the NMPC method, which was tested with manual controller switching.

\begin{table}
\vspace{5pt} 
\centering
\begin{threeparttable}
\caption{Position tracking performance}
\scriptsize 
\begin{tabular}{cccc}
\toprule  
Method & Max Pos Err (m) & Ave Pos Err (m) & Ave Vel Err (m/s)\\
\midrule  
NMPC (ideal)\textsuperscript{\textcolor{red}{*}} & $<0.1$ & $<0.1$ & $>0.3$ \\
Uniform PFTC\textsuperscript{\textcolor{blue}{\dag}} & $>6$ & $>0.5$ & / \\
Incremental PFTC\textsuperscript{\textcolor{green}{\ddag}} & $>1.5$ & $>0.3$ & / \\
Ours & $0.312$ & $0.100$ & $0.203$ \\
\bottomrule 
\end{tabular}
\begin{tablenotes}
\footnotesize
\item[\textcolor{red}{*}] Data extracted from Fig. 8 in \cite{nan2022nonlinear}, \textcolor{blue}{\dag} Data from Fig. 17 in \cite{ke2023uniform}, \textcolor{green}{\ddag} Data from Fig. 11 in \cite{beyer2023incremental}.
\end{tablenotes}
\end{threeparttable}
\label{table:experiment}
\end{table}

\begin{figure}
\vspace{5pt} 
    \centering
    \includegraphics[width=1\linewidth]{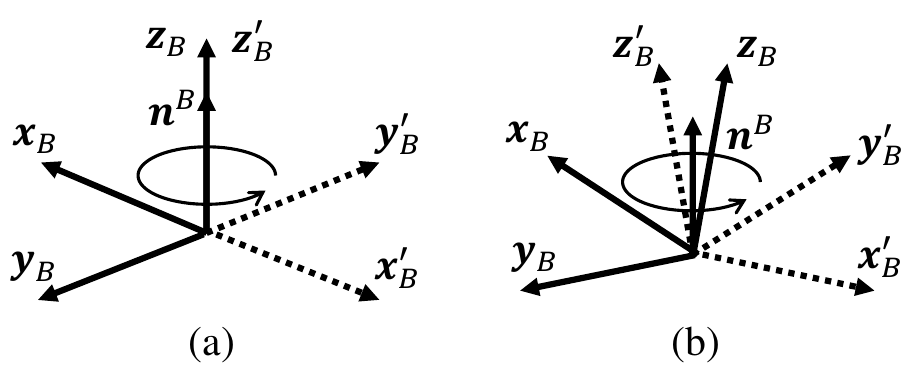}
    \caption{(a) The rotation axis $\boldsymbol{n}^B$ aligns with the body axis $\boldsymbol{z}_B$ $(\boldsymbol{n}^B = \begin{bmatrix} 0 & 0 & 1 \end{bmatrix}^T)$. (b) The rotation axis $\boldsymbol{n}^B$ does not align with the body axis $\boldsymbol{z}_B$ $(\boldsymbol{n}^B \neq \begin{bmatrix} 0 & 0 & 1 \end{bmatrix}^T)$.}
    \label{fig:primary_axis}
\end{figure}

We compute the primary axis $\boldsymbol{n}^B= \begin{bmatrix} n^B_x & n^B_y & n^B_z \end{bmatrix}^T =\boldsymbol{R}\boldsymbol{(q)}^T\begin{bmatrix} 0 & 0 & 1 \end{bmatrix}^T$, which represent the rotation direction in the body frame as shown in Fig.~\ref{fig:primary_axis}. 
As shown in the angular velocity curves and primary axis curves in Fig.~\ref{fig:real_0}, although we only penalize exceeding the IMU's measurement range in the rewards design, the policy autonomously controls the quadrotor to adjust the $\boldsymbol{n}^B$ axis to deviate from the $\boldsymbol{z}_B$ axis, as shown in Fig.\ref{fig:primary_axis}b. This adjustment redistributes part of the angular velocity to the $x$ and $y$ axes, effectively preventing the angular velocity along the $z$ axis from exceeding the measurement range. Notably, the direction of the $\boldsymbol{n}^B$ axis, which is handled automatically by our method, required manual configuration in previous work \cite{sun2018high}.

\section{CONCLUSIONS}
In this work, we propose a learning-based PFTC method for quadrotor capable of handling arbitrary single-rotor failures at any severity. The proposed Selector-Controller network structure effectively integrates the strengths of existing PFTC and AFTC approaches. By combining RL, BC and supervised learning, our method achieves significant performance improvements over state-of-the-art PFTC methods and leading AFTC methods in extensive simulations and real-world experiments. However, the current implementation has been validated in a motion capture environment with precise attitude and position measurements and low-speed flight tests. Future work should focus on enhancing its applicability to aggressive flight scenarios in outdoor environments.

\addtolength{\textheight}{-12cm}   





\bibliographystyle{IEEEtran}  

\end{document}